\documentclass[11pt,a4paper]{article}
\usepackage[hyperref]{acl2020}
\usepackage{times}
\usepackage{latexsym}
\usepackage{pgf-pie}
\usepackage{graphicx}
\usepackage{algorithm}
\usepackage{algpseudocode}

\usepackage[normalem]{ulem}
\useunder{\uline}{\ul}{}

\aclfinalcopy % Uncomment this line for the final submission
\def\aclpaperid{1297} %  Enter the acl Paper ID here

\title{ClarQ: A large-scale and diverse dataset for Clarification Question Generation}

\author{Vaibhav Kumar \\
  Language Technologies Institute \\
  Carnegie Mellon University \\
  \texttt{vaibhav2@cs.cmu.edu} \\\And
  Alan W Black \\
  Language Technologies Institute \\
  Carnegie Mellon University \\
  \texttt{awb@cs.cmu.edu} \\}

\date{}

\begin{document}
\maketitle
\begin{abstract}
Question answering and conversational systems are often baffled and need help clarifying certain ambiguities.
However, limitations of existing datasets hinder the development of large-scale models capable of generating and utilising clarification questions. 
In order to overcome these limitations, we devise a novel bootstrapping framework (based on self-supervision) that assists in the creation of a diverse, large-scale dataset of clarification questions based on post-comment tuples extracted from stackexchange. 
The framework utilises a neural network based architecture for classifying clarification questions. 
It is a two-step method where the first aims to increase the precision of the classifier and second aims to increase its recall.
We quantitatively demonstrate the utility of the newly created dataset by applying it to the downstream task of question-answering. The final dataset, ClarQ, consists of $\sim$2M examples distributed across 173 domains of stackexchange. 
We release this dataset\footnote{\url{https://github.com/vaibhav4595/ClarQ}} in order to foster research into the field of clarification question generation  with the larger goal of enhancing dialog and question answering systems.
 
\end{abstract}

\section{Introduction}
% The advent of voice based assistants like Alexa, Cortana, Siri etc. has made conversation one of the most convenient ways for natural languages interactions \citep{kiesel2018toward}. The use of conversations is not only limited to voice based assistants but is also quite prevalent on web-based forums like stackoverflow, where people indulge in multi-turn conversations for seeking answers to a particular question. Such 
The ubiquitous nature of conversations has led to the identification of various interesting problems \citep{clark2019makes}. 
% One such problem is the ability of a system to pose clarification questions \citep{rao2018learning, aliannejadi2019asking} when presented with a natural language question.
One of these problems is the ability of a system to ask for clarifications \citep{rao2018learning, aliannejadi2019asking} to a natural language question.

A user's complex information need is often lost due to the brevity of the posed question.
% When seeking answers, users often fail to specify their complex information needs in a single question.
This leads to an under-specified question containing information gaps which lowers the probability of providing the correct answer.
% This leads to an under-specified question containing information gaps which makes process of answer providing difficult. 
Thus, it would be an improvement if a conversational or a question answering system had a mechanism for refining user questions with follow-ups \citep{de2003analysis}. 
In literature, such questions have been termed \textbf{\textit{Clarification Questions}} \citep{de2003analysis, rao2018learning, rao2019answer}.

% Apart from resolving ambiguities within the question \citep{wang2018learning, rao2018learning, aliannejadi2019asking}, posing such questions can help conversation systems by driving the conversation deeper along with a better engagement of the user \citep{li2016learning, yu2016strategy}.

In the domain of question-answering, the major advantages of a clarification question are its ability to resolve ambiguities \cite{wang2018learning, aliannejadi2019asking} and to improve the probability of finding the most relevant answer.
For conversational systems, asking such questions help in driving the conversation deeper along with better engagement of the user \citep{li2016learning, yu2016strategy}.

% To this end, users often pose clarification questions on forums like stackexchange in diverse ways. 

% For example: sometimes a clarification question is as simple as demanding more information about particular entities within the posed question, whereas sometimes users posit clarification questions motivated by their personal experiences (or views) in order to understand the information demand of the posted question.

Recently, \citet{rao2018learning, rao2019answer} have provided a dataset based on stackexchange and used it for clarification question retrieval as well as generation. 
They also modify a dataset based on Amazon Question-Answering and Product Reviews \citep{mcauley2015image, mcauley2016addressing} to make it suitable for the same task. 
On the other hand, \citet{aliannejadi2019asking} created a dataset (Qulac) built on top of TREC web collections. 

However, there are several shortcomings to these datasets, which limit the development of generalizable and large-scale models aimed to tackle the problem of clarification question generation. 
The stackexchange dataset \citep{rao2018learning} is created by utilising simple heuristics. 
This adds a lot of noise, thereby reducing the number of actual clarification questions. 
It also limits the inclusion of diverse types of questions as it is collected from three similar domains (askubuntu, superuser and unix). 
The question generation model of \citet{rao2019answer} achieves a very low BLEU score when trained on this dataset. 
% This observation could confirm the presence of noise which precedes the model from generalization. 
On the other hand, the dataset based on Amazon reviews is a poor proxy for clarification questions because product descriptions are not actual questions demanding an answer and there is no information gap that needs to be addressed. 

% To overcome the shortcomings of these datasets, we come up with a more principled way to come up with a dataset which has (1) a high precision in terms of the actual clarification questions and (2) consists of clarifiation questions from a wide variety of domains leading to incorporation of diverse types of clarification question. We also quantitatively demonstrate the effectiveness of the collected clarification questions by utilsing them on the downstream task of question answer retrieval.

To overcome the shortcomings of existing datasets, we devise a novel bootstrapping framework based on self-supervision to obtain a dataset of clarification questions from various domains of stackexchange. 
The framework utilises a neural network based architecture to classify clarification questions. 
In a two step procedure, the framework first increases the precision of the classifier and then increases its recall. 
The first step is called down-sampling, where the classifier is iteratively trained on the most confident predictions (carried forward over from the previous iteration). 
The second step is the up-sampling procedure, where the classifier is iteratively trained by successively adding more positively classified examples. 
This step provides a boost in recall while restricting the drop in precision to a minimum. 
The classifier trained on the final iteration is then used for identification of clarification questions. 

The overall process ensures that the final dataset is less noisy and, at the same time, consists of a large and diverse number of examples. 
We must emphasize that, given the large amount of data available on stackexchange, a classifier with moderate recall still serves our purpose.
However, it is imperative that precision of the classifier be reasonably high.

%\section{Related Work}

\section{Methodology}
\href{https://stackexchange.com}{Stackexchange} is a network of online question answering websites. 
On these websites, users may comment on the original post with content such as third party URLs, clarifying questions, etc. 
We only want to select comments which act as clarifying questions and remove the rest as noise. 
To this end, we devise a bootstrapping framework for training a classifier capable of identifying clarifying questions.

The bootstrapping method utilises a neural network based classifier $\mathcal{L}$ which is posed with the task of clarification question detection. Formally, given a tuple $(p, q)$, where $p \in P$ is a post and $q \in q_p$ is a comment made on $p$, the task is to predict whether $q$ is an actual clarification question for $p$.
This makes it a binary classification problem, where a label $1$ indicates $q$ being an an actual clarification question and $0$ indicates otherwise.

\subsection{Data Collection}
We first utilise the stackexhange data dump available at \url{https://archive.org/details/stackexchange}. We extract the posts and the comments made by users on those posts from $173$ different domains. We remove all posts which did not have a provided answer. The comments made on the posts act as a potential candidate for clarifying question. This leads to 6,186,934 tuples of $(p, q)$.

\subsection{Bootstrapping}
First, we initialise a seed dataset that is used to train $\mathcal{L}$ using the process of iterative refinement as described later. Iterative-refinement itself is sub-divided into two parts: (1) Down-Sampling (2) Up-Sampling. 

% Given the large number of posts present on stackexchange, we can relax the condition of achieving a higher recall of classifying clarification questions and instead focus on obtaining a higher precision. This relaxation limits the amount of noise (or non-clarification questions) which might get added to the final dataset. At the same time, it also ensures the creation of a reasonably large dataset.
\subsubsection{Classifier $\mathcal{L}$}
We utilise a neural network based architecture for the classifier $\mathcal{L}$. Inspired by \citet{lowe2015ubuntu}, $\mathcal{L}$ utilises a dual encoder mechanism i.e it uses two separate LSTMs \citep{hochreiter1997long} for encoding a post $p$ and a question $q$. The dual encoder generates hidden representations $h_p$ and $h_{q}$ for $p$ and $q$ respectively. The resulting element-wise product of $h_p$ and $h_{q}$ is further passed on to fully connected layers before making predictions via softmax. More formally, the entire process can be summarised as follows:
\begin{equation}
    h_p = LSTM_{P}(p)
\end{equation}
\begin{equation}
    h_{q} = LSTM_{Q}(q)
\end{equation}
\begin{equation}
    h_{pq} = \phi(h_p \odot h_{q})
\end{equation}
\begin{equation}
    \hat{y} = Softmax(h_pq)
\end{equation}
where, $\odot$ represents the element-wise product, $\phi$ represents the non-linearity introduced by the fully connected layers and $\psi$ represents the final classification layer.

\subsubsection{Seed Selection}
\label{sec:seed}
In order to select seeds for the bootstrapping procedure, we consider all the collected posts but only use the last comment made on these posts as clarifying questions. We make the assumption that the comments act as a proxy for a clarification question. Later, we remove all $(p, q)$ tuples where $q$ does not have a question mark. 
Intuitively, the last comment can be a better signal for identifying clarifying questions as it has more chances of capsulizing the requirements of the original post. 
It can also be more opinionated than others. We then randomly sample a question from the same domain as that of the post and treat it as an instance of a negative clarification question. Thus each question gets paired with a positive and a negative clarification question. We denote this seed dataset as $D_0$. 

\begin{algorithm}[h]
	\caption{Iterative Refinement} 
	\label{algo}
	\begin{algorithmic}[1]

	\State $N \leftarrow 5$
	\State $D_0 \leftarrow$ Seed Data
% 	\State $\mathcal{L}_{list} \leftarrow$ []
	\State $\mathcal{T} \leftarrow$ Annotated Ground Truth
	\For {$i = 1, 2,\ldots,N$} \Comment{down-sampling}
	    \State $\mathcal{L} \leftarrow$ Classifier
	    \State train $\mathcal{L}$ on $D_{i - 1}$
	   % \State add $\mathcal{L}$ to $\mathcal{L}_{list}$
	    \State $D_{temp} \leftarrow$ []
	    \For {$(p, q) \in D_{i - 1}$}
	        \State $y \leftarrow \mathcal{L}(p, q)$
	        \If{ $y$ is true positive}
	            \State add $(p, q)$ to $D_{temp}$
	        \EndIf
	   \EndFor
	        \State Sort $D_{temp}$ using prediction confidence
	        \State $D_i \leftarrow$ top 40\% of $D_{temp}$
	        \State Randomly sample Negatives for $D_i$
	    
	\EndFor
	
	\State $S_N \leftarrow D_N$
	\For {$i = N, N - 1,\ldots 0$} \Comment{up-sampling}
	    \State $\mathcal{L} \leftarrow$ Classifier
	    \State train  $\mathcal{L}$ on $S_N$
	    \State $S_{temp} \leftarrow$ []
	    \For {$(p, q) \in D_{i - 1}$}
	        \State $y \leftarrow \mathcal{L}(p, q)$
	        \If{ $y$ is true positive}
	            \State add $(p, q)$ to $S_{temp}$
	        \EndIf
	    \EndFor
	    \State $S_{i - 1} \leftarrow S_{temp}$
	    \State Randomly Sample Negatives for $S_{i-1}$
	\EndFor
% 	\State train $\mathcal{L}$ on $S_N$
% 	\State $\mathcal{L}_{best} \leftarrow$ Learner
% 	\State $BestF1 \leftarrow 0$
% 	\For {$\mathcal{L} \in \mathcal{L}_{list}$} \Comment{choose best Learner}
% 	    \State $F1 \leftarrow$ F1 of $\mathcal{L}$ on $\mathcal{T}$
% 	    \If {$F1 > BestF1$}
% 	        \State $BestF1 \leftarrow F1$
% 	        \State $\mathcal{L}_{best} \leftarrow \mathcal{L}$
% 	   \EndIf
% 	\EndFor
    \State $\mathcal{L}_{best} \leftarrow$ Classifier
    \State $\mathcal{L}_{best}$ on $S_0$
	\State Use $\mathcal{L}_{best}$ to classify remaining data
	\end{algorithmic} 
\end{algorithm}

%\vspace{-1.5em}
\subsubsection{Iterative Refinement}
The procedure is described in Algorithm \ref{algo}. This entire process can be segmented into two parts.

% @@@@@@@@@@@@@@@@@
% Vk: Write about randomly sampled data from negative dataset. 
% @@@@@@@@@@@@@@@@@

\noindent \textbf{Down-Sampling}: The aim of this step is to increase the precision of the classifier. 
In the first iteration of this step, the classifier $\mathcal{L}$ is trained on the seed dataset $D_0$. 
After training is complete,  $\mathcal{L}$ classifies $D_0$ and the most confident $40\%$ of the positives are selected to train $\mathcal{L}$ in the next iteration. 
This process is continued for $N$ iterations. 
Each iteration leads to a new dataset $D_i$ (which is smaller in size than $D_{i - 1}$. 
Intuitively, the precision of $\mathcal{L}$ on the task of selecting actual clarification question should increase at the end of each iteration as it is successively trained only on the examples which it was more confident about in the previous round.

\noindent \textbf{Up-Sampling}:
This step is intended to improve the recall of $\mathcal{L}$ while restricting the loss of precision to a minimum. In the first iteration, $\mathcal{L}$ is trained on $S_N = D_N$ i.e the data obtained at the last iteration of the down-sampling procedure. After training is complete, $\mathcal{L}$ is used for classifying $D_{N-1}$ (which is obtained during the second-list iteration of the down-sampling process). The tuples which get classified as positive are used for training $\mathcal{L}$ in the next round. This process continued for $N$ iterations. Note that this procedure has two major differences to the iterative procedure of the down-sampling process. First, instead of using $\mathcal{L}$ for classifying the same dataset which it was trained on, it is used for classifying an up-sampled version of the current dataset. Second, it relaxes the condition of selecting $40\%$ of the most confident examples. Intuitively, this relaxation should help in increasing the recall of the classifier and at the same time should not drastically hamper the precision (as it operates only on the examples which it classifies as positives). 
% \noindent \textbf{Classifying Remaining Data}:

Note that, in order to provide the classifier with examples of non-clarifying questions, we randomly sample negative examples at the end of each iteration (during both up and down-sampling). This is similar to the way in which the $D_0$ is created.

%\vspace{-0.5em}
\subsubsection{Classifying Remaining Data}
At the end of the iterative refinement procedure, we obtain a dataset on which $\mathcal{L}$ can achieve a good precision and moderate recall on the task of classifying clarification questions. Thus, $\mathcal{L}$ is finally trained on $S_0$ and used for classifying the 6,186,934 tuples of $(p, q)$ extracted from stackexchange. We again emphasize that it is more important to obtain better precision, as it reduces the amount of noise added to the dataset. Given that there are a large number of $(p, q)$ tuples, a moderate recall can still ensure the incorporation of large and diverse types of $(p, q)$ tuples.

\section{Experimental Results}
%\vspace{-0.5em}
\begin{table}
\centering
\begin{tabular}{c|c|c|c}
\textbf{Iteration} & \textbf{Precision} & \textbf{Recall} & \textbf{F1} \\ \hline
1                  & 0.736              & 0.601           & 0.662       \\
2                  & 0.758              & 0.561           & 0.645       \\
3                  & 0.771              & 0.390           & 0.518       \\
4                  & 0.827              & 0.286           & 0.426       \\
5                  & 0.829              & 0.270           & 0.407      
\end{tabular}
\caption{Performance of the classifier on the annotated test set at the end of each iteration of the down-sampling procedure.}
\label{tab:down}
\end{table}

\begin{table}
\centering
\begin{tabular}{c|c|c|c}
\textbf{Iteration} & \textbf{Precision} & \textbf{Recall} & \textbf{F1} \\ \hline
1                  & 0.829              & 0.270           & 0.407       \\
2                  & 0.835              & 0.262           & 0.434       \\
3                  & 0.800                & 0.270           & 0.404       \\
4                  & 0.82               & 0.344           & 0.488       \\
5                  & 0.82               & 0.414           & 0.550      
\end{tabular}
\caption{Performance of the classifier on the annotated test set at the end of each iteration of the up-sampling procedure.}
\label{tab:up}
\end{table}

This section describes the results of the iterative refinement strategy.

\textbf{Test Set Creation}: We first create a manually annotated test set to evaluate the effectiveness of the classifier at each step of the iterative refinement process. For this, we randomly sample 100 $(p, q)$ tuples each from 7 different domains (Apple, cooking, gaming, money, photography, scifi, travel). These questions are either the last, second last or the third last comments of their corresponding posts. The annotated test set has a 7:3 ratio of positives to negatives.

% Initially there are a total of x 6million posts extracted from 174 different domains on stackexchange. Each posts consist of one or more comments. This leads to a total of this 6million tuples of post, question. Only using the last comment (which also contains a question mark) per post leads to a total of 900,000 examples.

\textbf{Seed Dataset}: It is created based on the method described in Section \ref{sec:seed}. It consists of 1,800,000 $(p, q)$ tuples, amongst which 50\% are randomly sampled negative instances. The classifier is then iteratively trained based on Algorithm \ref{algo}.

%\vspace{-0.3em}
\subsection{Results of Iterative Refinement}
The results of the down-sampling and the up-sampling procedure are discussed below:

\subsubsection{Down-Sampling}
Table \ref{tab:down} describes the performance of the classifier on the annotated test set during the down-sampling process. It can be clearly observed that the precision of the classifier increases with each iteration. Even though there is a substantial decline in recall, the down-sampling procedure helps in increasing the overall precision.

\subsubsection{Up-Sampling} 
Table \ref{tab:up} describes the performance of the classifier on the annotated test set during the up-sampling process. It can be clearly observed that recall of the classifier increases with each iteration, although the final recall (i.e at iteration 5) is lower than the recall obtained in the first iteration of the down-sampling process. Given that there are a large number of $(p, q)$ tuples, a drop in recall will not hamper the quality nor the diversity of the dataset. At the end of the process, we also observe that there is only a marginal drop in precision. Thus, at the end of the last iteration we are able to obtain a classifier which has a high precision and a reasonable recall.

\subsection{Downstream Utility}

We evaluate the utility of the clarification question in ClarQ by using it for the task of reranking answers. We first randomly sample 1000 $(p, q)$ tuples from 11 different domains (Apple, askubuntu, biology, cooking, english, gaming, money, puzzling, scifi, travel, unix). Corresponding to each tuple, we randomly sample a list of 99 answers (from the same domain as that of the post) and append the actual answer to this list. We first rerank the answers based on the post alone. Later, we rerank the answers by concatenating the post and the clarifying question. Based on the results from Table \ref{tab:downstream}, we observe that concatenating the clarification question to the post does help in improving the performance. The success of this experiment depicts the usefulness of our created dataset.

\begin{table}
\centering
\begin{tabular}{c|c|c}
\textbf{Metric} & \textbf{Without CQ} & \textbf{With CQ} \\ \hline
\textbf{P@1}    & 0.751               & 0.791            \\
\textbf{P@2}    & 0.399               & 0.416            \\
\textbf{P@3}    & 0.278               & 0.287            \\
\textbf{P@4}    & 0.214               & 0.220            \\ 
\textbf{P@5}    & 0.174               & 0.178            \\
\textbf{MRR}    & 0.791               & 0.816           
\end{tabular}
\caption{Performance on the task of question-answer retrieval. CQ stands for clarification question. P@k represents the precision at the kth position of the ranked list. MRR represents the Mean Reciprocal Rank.}
\label{tab:downstream}
\end{table}

% \begin{figure*}
% \setcounter{figure}{0}
% \centering
% \resizebox{\textwidth}{!}{%
% \begin{tikzpicture}
%   \pie[cloud, text=inside, after number, scale font, radius=8]{
% 21.78/math,
% 5.25/ru,
% 4.67/superuser,
% 4.11/tex,
% 3.42/electronics,
% 3.09/askubuntu,
% 3.04/serverfault,
% 2.63/unix,
% 2.54/pt,
% 2.4/english,
% 2.23/mathoverflow,
% 2.22/physics,
% 1.74/scifi,
% 1.73/softwareengineering,
% 1.72/es,
% 1.68/worldbuilding,
% 1.45/stats,
% 1.35/mathematica,
% 32.95/others}
% \end{tikzpicture}}
%  \caption{Distribution of the Clarifying Questions across different domains. The chart depicts the top 18 domains an increasing order from left to right. Rest of the domains are clubbed at the end of the spectrum in "others".}
%  \label{fig:dist}

% \end{figure*}

\begin{figure*}[t]
 \setcounter{figure}{0}
    \centering
    \includegraphics[width=\textwidth]{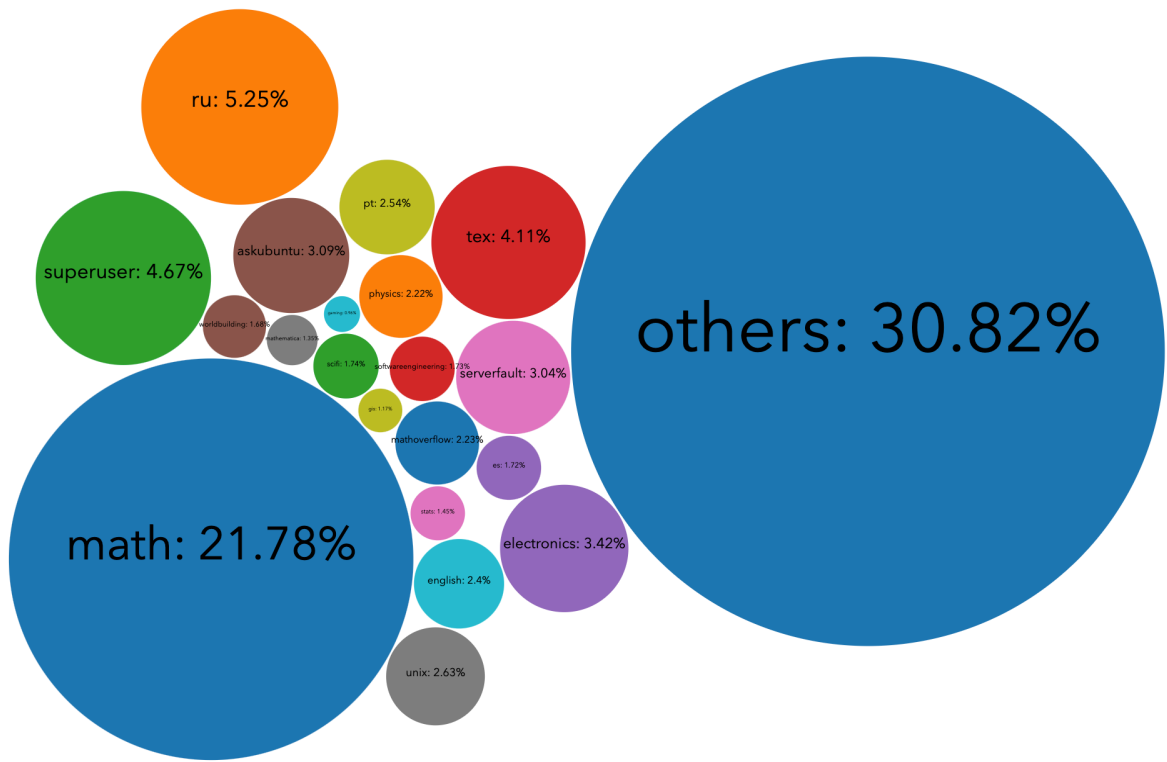}
    \caption{Distribution of the Clarifying Questions across different domains. The figure depicts the top 20 domains. Rest of the domains are clubbed at the end of the spectrum in "others".}
    \label{fig:dist}
\end{figure*}
\section{Dataset Statistics}

The classifier obtained at the end of iterative refinement procedure is used for classifying the initially collected $(p, q)$ tuples of 6,186,934. The classifier predicts 2,079,300 tuples as actual clarification questions. As can be seen from Figure \ref{fig:dist}, these tuples are unequally distributed across 173 different domains. The top 20 domains account for $69.18\%$ of the total $(p, q)$ tuples in the dataset. The remaining 155 domains account for the remaining $30.82\%$ of the total number of tuples. 

It is noteworthy that our provided dataset also comprises of actual answers to each post. This would help researchers in evaluating the quality of the clarification questions in a standalone perspective and at the same time with respect to the downstream task of question-answering.

\section{Conclusion and Future Work}
In this paper, we present a diverse, large-scale dataset (\textbf{ClarQ}) for the task of clarification question generation.
It is created by a two-step iterative bootstrapping framework based on self-supervision.
ClarQ consists of $\sim$2M post-question tuples spanning 173 different domains. 
We hope that this dataset will encourage research into clarification question generation and, in the long run, enhance dialog and question-answering systems.

%%%RESULTS SECTION%%%%t

\section*{Acknowledgments}
We would like to extend our sincere gratitude to Abhimanshu Mishra, Mrinal Dhar and Yash Kumar Lal for helping us understand the structure of the comments and their distribution across domains.

\bibliography{acl2020}
\bibliographystyle{acl_natbib}

\appendix

\end{document}